\documentclass{article}

\usepackage[preprint, nonatbib]{nips_2018}

\usepackage{float}

\usepackage[square,numbers]{natbib}
\usepackage[utf8]{inputenc} 
\usepackage[T1]{fontenc}    
\usepackage{hyperref}       
\usepackage{url}            
\usepackage{booktabs}       
\usepackage{amsfonts}       
\usepackage{nicefrac}       
\usepackage{microtype}      
\usepackage{graphicx}
\usepackage{tabularx}
\usepackage{hyperref}

\usepackage{graphics} 
\usepackage{epsfig} 
\usepackage{mathptmx} 
\usepackage{times} 
\usepackage{amsmath} 
\usepackage{amssymb}
\usepackage{subcaption}
\usepackage{algorithmic} 
\usepackage{multirow} 
\usepackage{float}
\usepackage{array}
\usepackage{booktabs}
\usepackage[linesnumbered,boxed,ruled,commentsnumbered]{algorithm2e}

\usepackage{xcolor}
\usepackage{soul, color}
\usepackage{colortbl}
\definecolor{myred}{RGB}{255,228,225}
\definecolor{mygreen}{RGB}{220,255,220}

\title{DiffYOLO: Object Detection for Anti-Noise via YOLO and Diffusion Models}

\author{
  Yichen Liu \\
  \texttt{liuyichen21@mails.ucas.ac.cn} \\
\And
 Huajian Zhang \\
  \texttt{zhanghj@impcas.ac.cn} \\
\And
 Daqing Gao \\
  \texttt{gaodq@impcas.ac.cn} \\
}

\begin{document}
\maketitle

\begin{abstract}
	Object detection models represented by YOLO series have been widely used and have achieved great results on the high quality datasets, but not all the working conditions are ideal. To settle down the problem of locating targets on low quality datasets, the existing methods either train a new object detection network, or need a large collection of low-quality datasets to train. However, we propose a framework in this paper and apply it on the YOLO models called DiffYOLO. Specifically, we extract feature maps from the denoising diffusion probabilistic models to enhance the well-trained models, which allows us fine-tune YOLO on high-quality datasets and test on low-quality datasets. The results proved this framework can not only prove the performance on noisy datasets, but also prove the detection results on high-quality test datasets. We will supplement more experiments later (with various datasets and network architectures).
\end{abstract}
\section{Introduction}
\label{sec:intro}

YOLO has become prevailed in target detection tasks, from automatic driving to medical image processing. Alice Froidevaux et al. used YOLO to detect vehicles through satellite images\cite{froidevaux2020vehicle}; Sudipto Paul et al. applied YOLO to brain cancer recognition on MRI images\cite{paul2022brain}; Ethan Grooby et al. explored automated facial landmark detection using YOLO\cite{grooby2023neonatal}. Although YOLO has achieved great success in object detection tasks, capturing objects from images with noises is still a great challenge. Normally object detection models are trained on high quality images, but the test condition may not be so ideal. Fig.1 shows on the test images with noise, a well-trained YOLO on high quality datasets has poor detection results. If these models trained on high-quality data sets can perform well on noise test sets with simple enhancements, then the trained models can be better utilized.

Transfer learning on pretrained models is an important method to make full use of pre-trained models. It first appeared in language models called fine-tune\cite{houlsby2019parameter}, bringing many benefits, such as making training more efficiently and less dependent on high-quality training sets, therefore we hope to find a method to  leverage other well-trained models to improve the performance of YOLO models.

Denoising diffusion probabilistic models(DDPM) was put forward by Sohl-Dickstein et al., has shown great advantage in many generation tasks\cite{sohldickstein2015deep,ho2020denoising}. Othmane Laousy et al. demonstrated that the diffusion method is not susceptible to perturbations \cite{laousy2023towards}, so we decided to incorporate the diffusion model into the YOLO model.

Therefore, we propose a framework in this paper for improving the noise resistance of models already trained on high-quality data sets, called DiffYOLO. We first extract some features from the Unet of the already trained Diffusion models, fuse them, and then splice them into the neck module of YOLO. The feature extracted by such a diffusion model can improve the YOLO model to obtain the anti-noise ability of Unet. Figure 1 shows our proposed framework compared to baseline's test results.

\begin{figure}[H]
\centering
\subcaptionbox{\label{1}}{\includegraphics[width = .48\linewidth]{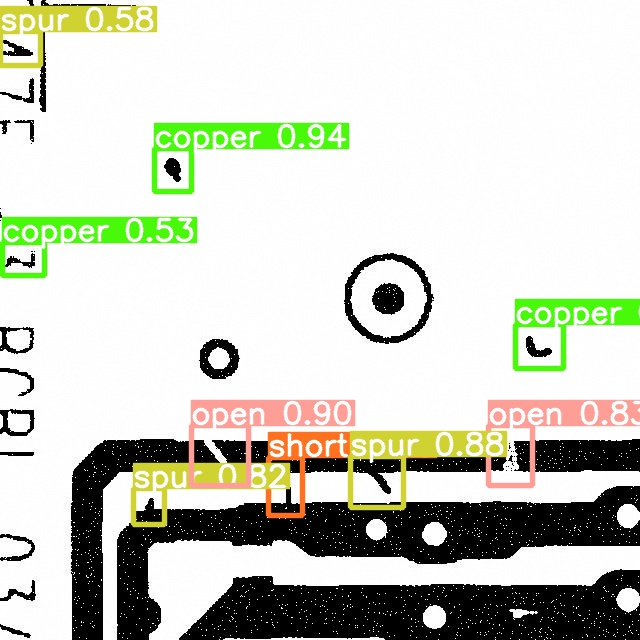}}\hfill
\subcaptionbox{\label{2}}{\includegraphics[width = .48\linewidth]{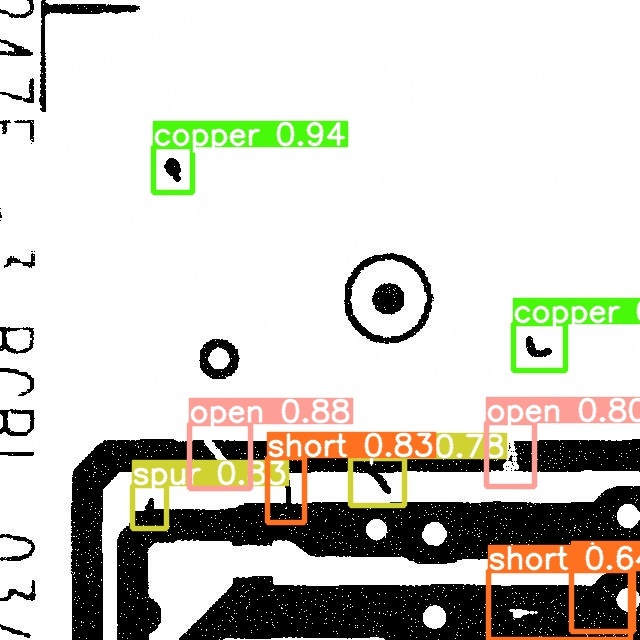}}
\caption{(a)Defect detection results by YOLOv5 on the image with noise; (b)Defect detection results by DiffYOLO on the image with noise}
\label{demo}
\end{figure}

\begin{figure*}[htbp]
\centerline{\includegraphics[width=0.78\textwidth]{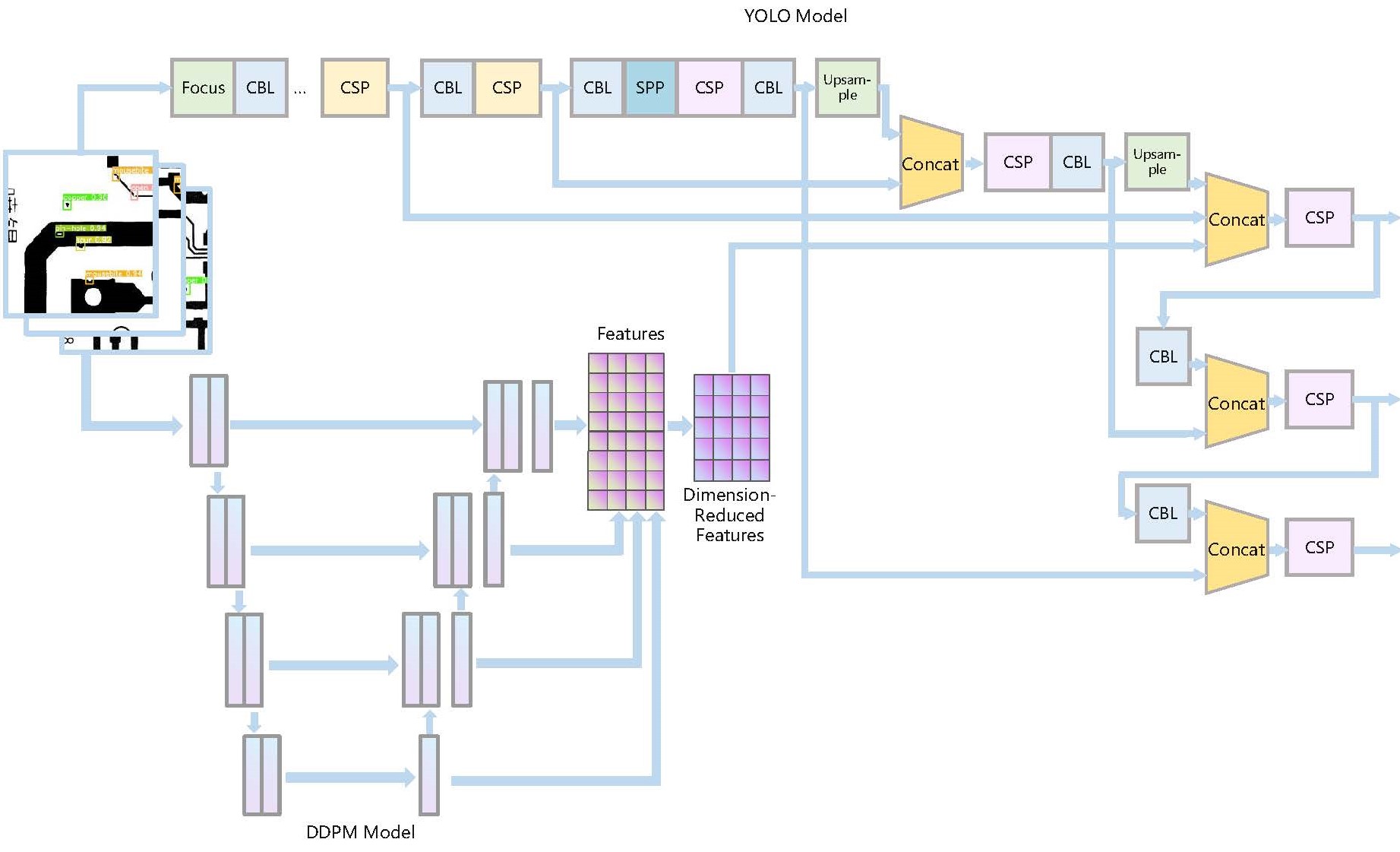}}
\caption{The overall framework of DiffYolo.}
\label{flow}
\end{figure*}

The contributions of our paper are:
\begin{itemize}
\item First, we build our work upon Dmitry Baranchuk et al. as the fundamental to propose for the first time, to use diffusion models to improve the YOLO detection methods. From a series of methods we demonstrate that extracting features from the pretrained diffusion models and injecting them to a traditional target detection model will make a great progress in anti-noise applications.
\item Second, our method is to finetune the pretrained models instead of training by ourselves, allowing us to use less resources to reach a higher accuracy.
\item Third, we tested our model separately on images with different levels of noise, and the experimental results showed that our framework performed better even with the original image without noise.
\end{itemize}

\section{Related work}
\label{sec:format}
\subsection{Object Detection}

Object detection is a typical task in computer vision, and many models were proposed to settle down this problem. Ross Girshick et al. proposed R-CNN to extract features using convolutional models after selecting candidate boxes\cite{girshick2014rich}. Faster RCNN was proposed then by Ross Girshick et al. to improve the detection speed\cite{girshick2015fast}. These methods are two-stage detection methods, and another common detection method category is the one-stage methods such as the YOLO series. From YOLOv1\cite{redmon2016look} to YOLOx\cite{ge2021yolox} and PP-YOLOE\cite{xu2022ppyoloe}, YOLO model series evolved to be more accurate and faster. The baseline model in this paper is the classic one-stage detector YOLOv5, and we improved its performance in noisy environments.

\subsection{Diffusion Models}
The target of diffusion models is to reconstruct data from random noises. Different from former reconstruct models like GAN, diffusion models generate target distribution step by step, from which each step is modeled by a deep neural network to build a denoising model. That is to say, diffusion models learn a series of state transitions to transform noise. Diffusion models on image or video segmentation\cite{amit2021segdiff}, image colorization (Song et al., 2021\cite{staal2004ridge}) and anomaly detection \cite{mousakhan2023anomaly}. In our work, we prove diffusion model can improve the noise resistance of other networks.

\subsection{Anti-Noise}
While it is easy to get pre-trained models (most open-source models are already pretrained with high quality datasets), it is not always possible to get clear images for actual object detection. For example, when the image to be detected in the industrial scene is transmitted to the computer, the transmission interference will bring difficulties to the target detection. Images captured in foggy or low-light weather are also subject to interference, which is a challenge for object detection in the field of autonomous driving.
NoisyNet use reinforcement learning methods to add noise into networks to improve performance\cite{fortunato2017noisy}. Wenyu Liu et al. proposed IA-YOLO model\cite{liu2022image} in which each image can be adaptively enhanced for better detection performance. DANNet model is based on GAN to segment night images \cite{wu2021dannet}.

\section{Method}
\label{sec:method}

In this section, we describe our framework DiffYOLO, an improved YOLO model. Our experiments prove that for baseline YOLOv5 model, noise will seriously interfere with the detection of the model. In order to deal with the challenge of image interference, we propose a framework of object detection, which can improve the anti-interference ability of the model. The framework can also be applied to other models that need to resist noise.

We first describe the process of extracting features from DDPM. Diffusion models are divided into the forward process of adding noise and the backward process of removing noise. The forward process can be regarded as a Markov process, where the image $x_t$ at the current point in time is only related to the image $x_{t-1}$ at the previous moment:
\begin{gather}
q(x_t|x_{t-1})=N(x_t,\sqrt{1-\beta_t}x_{t-1},\beta_tI)
\end{gather}

After re-parameterized sampling, we can get the distribution of image $x_t$ at any time $t$, and only related to $x_0$:
\begin{gather}
q(x_t|x_0)=N(x_t;\sqrt{\overline{\alpha}_t }x_{0},(1-\overline{\alpha}_t)I)
\end{gather}
from which$\epsilon \sim N(0,1), {\alpha}_t=1-\beta_t, \overline{\alpha}_t= {\textstyle \prod_{T}^{i=1}{\alpha}_i} $.

The reverse process is the de-noising process of the final noise image, that is, the original image is restored by random noise $x_T$. The diffusion model learns not the distribution, but the noise at time $t$. According to the re-parameterization technique, the final diffusion model can de-noise and return $x_0$.

In our experiment we use model proposed by (Dhariwal \& Nichol, 2021), extract its features at different levels (as shown in the Fig.2), and add them in a unified dimension. Generate a correction variable for a model trained to enhance noise resistance. Since it is extracted from the diffusion model, we think that this feature carries information on how to resist the noise, and we expect the model to learn this information through its input. We input the feature into the tail of the first branch of the neck module of YOLOv5, which not only retains the original image information, but also adds the anti-noise part.

In our proposed framework, the model to be enhanced does not need to be retrained, because we only correct the feature extracted by diffusion for the intermediate result of a certain part of the model. This not only greatly saves training time, but also can be applied to different pre-training models.

\section{Experiments}
\label{sec:exp}

We tested our approach on the PCB defect dataset(Deep PCB) and compared it with baseline YOLOv5.
\subsection{Dataset}
DeepPCB is a real data set of 1500 samples obtained by a linear scan CCD, which contains six common PCB defects: open, short, mouse bite, spur, copper, and pin-hole. To simulate the noise that might occur in real industrial scenarios, we trained our model with high quality datasets using our framework, and used zero noise, Gaussian noise, Salt and Pepper nois and Possion noise to separately test our model and the baseline model. 


\subsection{Experiment Results}

In actual operation, in order to run efficiently, we disabled the load mosaic module of YOLOv5 during train. The method we adopt is to store the features in the disk in advance and load them into the model when the features are needed, instead of generating the in real time (another feasible way is to enable the load mosaic module and carry out feature calculating and training after the load mosaic module). The results of different datasets are shown in Table.\ref{tabnormal},\ref{tabgaussian},\ref{tabspeckle},\ref{tabpoison}.

\begin{table}[h]
\centering
\caption{Detection Results of High Quality Datasets}
\label{tabnormal}
\begin{subtable}[t]{0.6\linewidth}
    \caption{Detection Results of Yolov5 Model}
\begin{tabular}{lcccc}
\toprule
\textbf{Class}                     & \textbf{P}    & \textbf{R}  & \textbf{mAP@0.5}    & \textbf{mAP@0.95}   \\
\midrule

all & 0.9777 & 0.954 & \cellcolor{myred}0.971 &\cellcolor{myred}0.937\\
open & 0.976 & 0.982 & \cellcolor{myred}0.977 &\cellcolor{myred}0.959\\
short & 0.959 & 0.892 & \cellcolor{myred}0.937 &\cellcolor{myred}0.916\\
mouse bite & 0.982 & 0.948 & \cellcolor{myred}0.973 &\cellcolor{myred}0.935\\
spur & 0.985 & 0.963 & \cellcolor{myred}0.974 &\cellcolor{myred}0.901\\
copper & 0.993 & 0.985 & \cellcolor{myred}0.992 &\cellcolor{myred}0.951\\
pin-hole & 0.97 & 0.956 & \cellcolor{myred}0.977 &\cellcolor{myred}0.914\\
\bottomrule
\end{tabular}
\end{subtable}

\begin{subtable}[t]{0.6\linewidth}
    \caption{Detection Results of DiffYolo Model}
\begin{tabular}{lcccc}
\toprule
\textbf{Class}                     & \textbf{P}    & \textbf{R}  & \textbf{mAP@0.5}    & \textbf{mAP@0.95}   \\
\midrule

all & 0.984 & 0.971 & \cellcolor{mygreen}0.982 &\cellcolor{mygreen}0.978\\
open & 0.976 & 0.994 & \cellcolor{mygreen}0.988 &\cellcolor{mygreen}0.970\\
short & 0.984 & 0.954 & \cellcolor{mygreen}0.975 &\cellcolor{mygreen}0.932\\
mouse bite & 0.976 & 0.959 & \cellcolor{mygreen}0.979 &\cellcolor{mygreen}0.985\\
spur & 0.978 & 0.978 & \cellcolor{mygreen}0.980 &\cellcolor{mygreen}0.926\\
copper & 1.000 & 0.993 & \cellcolor{mygreen}0.995 &\cellcolor{mygreen}0.957\\
pin-hole & 0.992 & 0.949 & \cellcolor{mygreen}0.974 &\cellcolor{mygreen}0.947\\
\bottomrule
\end{tabular}
\end{subtable}
\end{table}
\begin{table}[h]
\centering
\caption{Detection Results under Guassian Noise}
\label{tabgaussian}
\begin{subtable}[t]{0.6\linewidth}
    \caption{Detection Results of Yolov5 Model}
\begin{tabular}{lcccc}
\toprule
\textbf{Class}                     & \textbf{P}    & \textbf{R}  & \textbf{mAP@0.5}    & \textbf{mAP@0.95}   \\
\midrule

all & 0.824 & 0.642 & \cellcolor{myred}0.751 &\cellcolor{myred}0.515\\
open & 0.94 & 0.855 & \cellcolor{myred}0.91 &\cellcolor{myred}0.591\\
short & 0.641 & 0.823 & \cellcolor{myred}0.803 &\cellcolor{myred}0.496\\
mouse bite & 0.981 & 0.308 & \cellcolor{myred}0.647 &\cellcolor{myred}0.458\\
spur & 0.964 & 0.787 & \cellcolor{mygreen}0.884 &\cellcolor{myred}0.572\\
copper & 0.963 & 0.970 &\cellcolor{myred} 0.984 &\cellcolor{mygreen}0.776\\
pin-hole & 0.455 & 0.11 & \cellcolor{myred}0.276 &\cellcolor{myred}0.198\\
\bottomrule
\end{tabular}
\end{subtable}

\begin{subtable}[t]{0.6\linewidth}
    \caption{Detection Results of DiffYolo Model}
\begin{tabular}{lcccc}
\toprule
\textbf{Class}                     & \textbf{P}    & \textbf{R}  & \textbf{mAP@0.5}    & \textbf{mAP@0.95}   \\
\midrule

all & 0.832 & 0.678 & \cellcolor{mygreen}0.775 &\cellcolor{mygreen}0.582\\
open & 0.972 & 0.842 & \cellcolor{mygreen}0.913 &\cellcolor{mygreen}0.666\\
short & 0.736 & 0.900 & \cellcolor{mygreen}0.902 &\cellcolor{mygreen}0.653\\
mouse bite & 0.976 & 0.483 & \cellcolor{mygreen}0.732 &\cellcolor{mygreen}0.574\\
spur & 0.960 & 0.706 & \cellcolor{myred}0.845 &\cellcolor{mygreen}0.603\\
copper & 0.964 & 0.978 & \cellcolor{mygreen}0.987 &\cellcolor{myred}0.728\\
pin-hole & 0.386 & 0.162 & \cellcolor{mygreen}0.280 &\cellcolor{mygreen}0.270\\
\bottomrule
\end{tabular}
\end{subtable}
\end{table}

Although the baseline YOLO's performance and our DiffYOLO performance are all significantly degraded by certain types of noise (e.g., Gauss) impact, our experiments still show that our method outperforms the baseline method in most categories under Gaussian, Salt and Pepper, and Possion noise, that is to say, the procedure of extracting features from diffision model can help baseline models acquire noise resistance. In addition to improving the noise resistance of the YOLO model, our approach can also improve the target detection results of the model itself against high-quality test datasets. In other words, our framework can improve the performance of the baseline model for target detection as a whole.


\begin{table}[h]
\centering
\caption{Detection Results under Salt and Pepper Noise}
\label{tabspeckle}
\begin{subtable}[t]{0.6\linewidth}
    \caption{Detection Results of Yolov5 Model}
\begin{tabular}{lcccc}
\toprule
\textbf{Class}                     & \textbf{P}    & \textbf{R}  & \textbf{mAP@0.5}    & \textbf{mAP@0.95}   \\
\midrule

all & 0.782 & 0.478 & \cellcolor{myred}0.655 &\cellcolor{myred}0.446\\
open & 0.920 & 0.632 & \cellcolor{myred}0.791 &\cellcolor{mygreen}0.512\\
short & 0.595 & 0.586 & \cellcolor{myred}0.648 &\cellcolor{myred}0.389\\
mouse bite & 1 & 0.100 & \cellcolor{myred}0.550 &\cellcolor{mygreen}0.414\\
spur & 0.987 & 0.556 & \cellcolor{mygreen}0.775 &\cellcolor{myred}0.519\\
copper & 0.814 & 0.970 & \cellcolor{myred}0.965 &\cellcolor{mygreen}0.745\\
pin-hole & 0.375 & 0.02 & \cellcolor{myred}0.199 &\cellcolor{myred}0.100\\
\bottomrule
\end{tabular}
\end{subtable}

\begin{subtable}[t]{0.6\linewidth}
    \caption{Detection Results of DiffYolo Model}
\begin{tabular}{lcccc}
\toprule
\textbf{Class}                     & \textbf{P}    & \textbf{R}  & \textbf{mAP@0.5}    & \textbf{mAP@0.95}   \\
\midrule

all & 0.848 & 0.502 & \cellcolor{mygreen}0.698 &\cellcolor{mygreen}0.450\\
open & 0.947 & 0.648 & \cellcolor{mygreen}0.809 &\cellcolor{myred}0.508\\
short & 0.792 & 0.731 & \cellcolor{mygreen}0.799 &\cellcolor{mygreen}0.494 \\
mouse bite & 1 & 0.123 & \cellcolor{mygreen}0.561 &\cellcolor{myred}0.379 \\
spur & 0.971 & 0.504 &\cellcolor{myred} 0.741 &\cellcolor{mygreen}0.482\\
copper & 0.805 & 0.978 & \cellcolor{mygreen}0.970 &\cellcolor{myred}0.650\\
pin-hole & 0.571 & 0.030 & \cellcolor{mygreen}0.305 &\cellcolor{mygreen}0.188\\
\bottomrule
\end{tabular}
\end{subtable}
\end{table}

\begin{table}[h]
\centering
\caption{Detection Results under Possion Noise}
\label{tabpoison}
\begin{subtable}[t]{0.6\linewidth}
    \caption{Detection Results of Yolov5 Model}
\begin{tabular}{lcccc}
\toprule
\textbf{Class}                     & \textbf{P}    & \textbf{R}  & \textbf{mAP@0.5}    & \textbf{mAP@0.95}   \\
\midrule

all & 0.977 & 0.956 & \cellcolor{myred}0.972 &\cellcolor{myred}0.737\\
open & 0.976 & 0.982 & \cellcolor{myred}0.977 &\cellcolor{myred}0.659\\
short & 0.959 & 0.900 & \cellcolor{myred}0.941 &\cellcolor{myred}0.664 \\
mouse bite & 0.982 & 0.948 & \cellcolor{myred}0.561 &\cellcolor{myred}0.735 \\
spur & 0.985 & 0.963 & \cellcolor{myred}0.974 &\cellcolor{myred}0.699\\
copper & 0.993 & 0.985 & \cellcolor{myred}0.992 &\cellcolor{myred}0.851\\
pin-hole & 0.970 & 0.956 & \cellcolor{myred}0.977 &\cellcolor{myred}0.814\\
\bottomrule
\end{tabular}
\end{subtable}

\begin{subtable}[t]{0.6\linewidth}
    \caption{Detection Results of DiffYolo Model}
\begin{tabular}{lcccc}
\toprule
\textbf{Class}                     & \textbf{P}    & \textbf{R}  & \textbf{mAP@0.5}    & \textbf{mAP@0.95}   \\
\midrule

all & 0.984 & 0.970 & \cellcolor{mygreen}0.981 &\cellcolor{mygreen}0.777\\
open & 0.976 & 0.994 & \cellcolor{mygreen}0.988 &\cellcolor{mygreen}0.720\\
short & 0.984 & 0.954 & \cellcolor{mygreen}0.975 &\cellcolor{mygreen}0.732\\
mouse bite & 0.976 & 0.978 & \cellcolor{mygreen}0.979 &\cellcolor{mygreen}0.787\\
spur & 0.978 & 0.978 & \cellcolor{mygreen}0.980 &\cellcolor{mygreen}0.725\\
copper & 1 & 0.993 & \cellcolor{mygreen}0.995 &\cellcolor{mygreen}0.857\\
pin-hole & 0.992 & 0.941 & \cellcolor{mygreen}0.978 &\cellcolor{mygreen}0.844\\
\bottomrule
\end{tabular}
\end{subtable}
\end{table}

\section{Conclusion}
\label{sec:conclusion}

In this paper, denoising diffusion probabilistic models are used, from which feature maps are extracted to improve the anti-noise capability of baseline model. The experiments in this paper successfully prove that the learning results of the diffusion model can be extracted and used as the representation learner of the target detection problem.  In contrast to previous methods such as the direct improvement of YOLO, we propose a framework that can be applied to different large models that have been trained (rather than being limited to one specific model). The advantage of our framework is that it can reuse pretrained models and improve the results of high-quality test datasets without noise. A significant limitation of DiffYOLO is that many industrial IoT devices that may require this capability do not have the computational resources themselves to extract feature maps from large models and then retrain them, yet the data sets they produce may be constantly changing, i.e., migrating. Over time, models may fail as data characteristics migrate. However, we believe that simplification of the model will solve this problem in the future, making the framework more portable and easier to train.

\vfill\pagebreak

\bibliography{ref.bib}

\end{document}